\documentclass[10pt,twocolumn,letterpaper]{article}

\usepackage[pagenumbers]{cvpr} 

\newcommand{\ode}{ODE--ViT }

\usepackage{microtype}

\usepackage{tikz}
\usetikzlibrary{3d,arrows.meta,backgrounds,decorations.pathmorphing}
\usetikzlibrary{intersections,shapes.arrows}

\usepackage{xspace}
\usepackage{amsthm}
\usepackage{graphics}
\usepackage{booktabs}
\usepackage{multirow}
\usepackage{graphicx}
\usepackage[table,xcdraw]{xcolor}

\newtheorem{prop}{Proposition}
\newtheorem{obs}{Observation}
\usepackage{dblfloatfix}

\usepackage{soul}
\setuldepth{foobar}

\definecolor{cvprblue}{rgb}{0.21,0.49,0.74}
\usepackage[pagebackref,breaklinks,colorlinks,allcolors=cvprblue]{hyperref}

\title{ODE-ViT: Plug \& Play Attention Layer from the Generalization of the ViT as an Ordinary Differential Equation. }

\author{
Carlos Boned Riera\\
{\small Computer Vision Center (CVC)} \\
{\small Universitat Autònoma de Barcelona} \\
{\tt\small cboned@cvc.uab.cat}
\and
David Romero Sánchez \\
{\small Mathematical Research Center (CRM)} \\
{\small Universitat Autònoma de Barcelona} \\
{\tt\small dromero@crm.cat}
\and
Oriol Ramos Terrades \\
{\small Computer Vision Center (CVC)} \\
{\small Universitat Autònoma de Barcelona} \\
{\tt\small oriolrt@cvc.uab.cat}
}

\begin{document}
\twocolumn[{%
\renewcommand\twocolumn[1][]{#1}%
\maketitle
\centering
\captionsetup{type=figure}
    \centering
    \includegraphics[width=\textwidth]{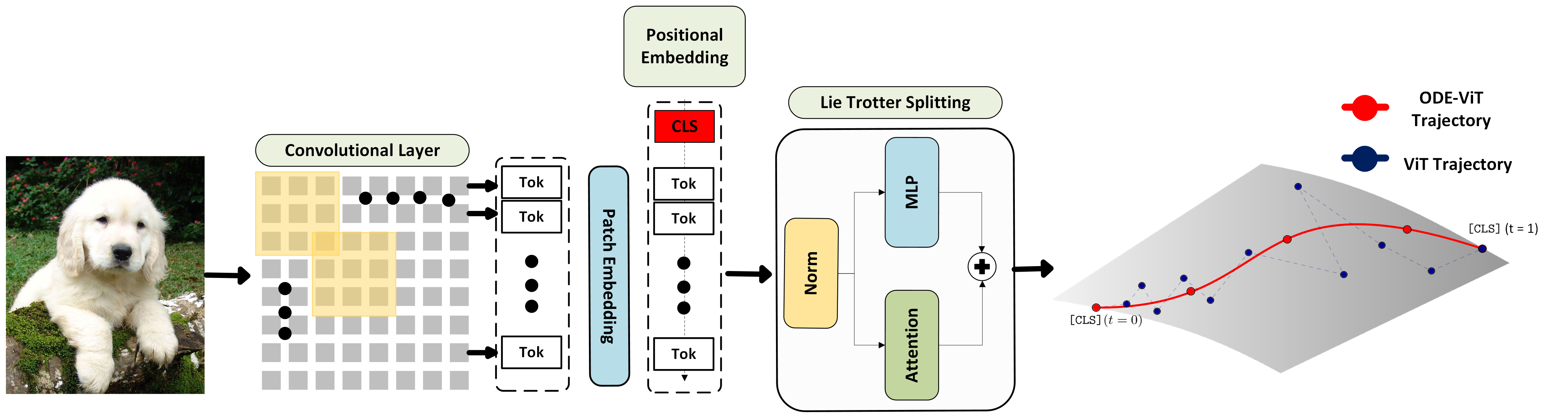}
\vspace{-8mm}
\caption{Overall architecture used in this work. The input image is processed through a convolutional layer to extract patch tokens. These tokens are mapped into $ d$-dimensional space, where the attention block generates the continuous trajectory. In red, there is the \texttt{[CLS]} trajectory generated by evaluating $\psi(x(t), t;\theta)$ $N$ times, and in blue the \texttt{[CLS]} trajectory generated by a 12-layer ViT.} 
\label{fig:odeVit}
\vspace{3mm}
 
}]


\begin{abstract}

In recent years, increasingly large models have achieved outstanding performance across CV tasks.  
However, these models demand substantial computational resources and storage, and their growing complexity limits our understanding of how they make decisions.  
Most of these architectures rely on the attention mechanism within Transformer-based designs.  
Building upon the connection between residual neural networks and ordinary differential equations (ODEs), we introduce \textbf{ODE-ViT}, a Vision Transformer reformulated as an ODE system that satisfies the conditions for well-posed and stable dynamics.  
Experiments on CIFAR-10 and CIFAR-100 demonstrate that \textbf{ODE-ViT} achieves stable, interpretable, and competitive performance with up to one order of magnitude fewer parameters, surpassing prior ODE-based Transformer approaches in classification tasks.  
We further propose a \textbf{plug-and-play} teacher–student framework in which a discrete ViT guides the continuous trajectory of \textbf{ODE-ViT} by treating the teacher’s intermediate representations as solutions of the ODE.  
This strategy improves performance by more than 10\% compared to training a free ODE-ViT from scratch.

\end{abstract}    
\section{Introduction}
\label{sec:intro}
With the emergence of very large models, the fields of computer vision (CV) and natural language processing (NLP) have been revolutionized.  
While relatively compact architectures such as ResNet or BERT can be trained with modest hardware resources, modern large language models (LLMs) and vision–language models (VLMs) have made both training and inference increasingly demanding in terms of computation and memory usage.  
Furthermore, as model complexity grows, the mechanisms underlying their decision-making processes become more opaque.  
This raises a natural question: \textit{can we design more compact and interpretable architectures without sacrificing performance?}

Many works have explored model compression through \textbf{knowledge distillation}~\cite{fan2024scalekd,KD_survey}, where a lightweight network (student) is trained under the supervision of a larger model (teacher).  
Other approaches focus on reducing parameter counts through techniques such as pruning~\cite{Brain_Damage}, low-rank adaptation (LoRA)~\cite{hu2022lora}, and quantization~\cite{Jacob_2018_CVPR}.  
Although these methods achieve impressive efficiency gains, they generally do not enhance interpretability or shed light on the internal decision mechanisms of the resulting models.

Recent advances in \textbf{reasoning models}~\cite{deepseekai2025deepseekr1,chain_of_though} have sought to improve interpretability by explicitly structuring the reasoning process during training or inference.  
However, these models often lack compactness and require extensive computational resources, leaving open the challenge of achieving both interpretability and efficiency within a unified framework.

A theoretical direction reinterprets the residual networks as an ordinary differential equation (ODE). They show that the residual updates can be interpreted as discretized steps of a continuous dynamical system. Theoretical works have established formal connections between residual networks and ODEs.~\cite{conf/nips/SanderAP22,teshima2020universalapproximationpropertyneural,li2020deeplearningdynamicalsystems} 

A ResNet~\cite{he2016residual,he2016identity} of depth $N$ evolves from an initial point $x_0 \in \mathbb{R}^d$ according to
\[
x_{n+1} = x_n + f(x_n, \theta^N_n),
\]
And by scaling the residual update, we obtain the following.
\begin{equation}\label{eq:residual_equation}
    x_{n+1} = x_n + \frac{1}{N} f(x_n, \theta^N_n),
\end{equation}
which approximates the continuous ordinary differential equation (ODE)
\begin{equation}\label{eq:dynamic_equation}
    \dot{x}(t) = \psi(x(t), t; \theta), \quad x(0) = x_0,
\end{equation}
as $N \to \infty$.

This equivalence enables different theoretical tools of dynamical systems theory that allow us to analyze and interpret the behavior of these models. In fact, the Picard–Lindelöf theorem gives a set of conditions under which an initial value problem has a unique solution.
One of the conditions of the theorem is that the ODE need to be Lipschitz continuous. Being Lipschitz continuous is important for improving training stability, enabling faster convergence, and enhancing model robustness and generalization. However, not all models are Lipschitz continuous. Transformers are not Lipschitz\cite{kim2021lipschitzconstantselfattention} and several works~\cite{qi2023lipsformerintroducinglipschitzcontinuity,castin2024smoothattention,hu2024specformer,kim2021lipschitzconstantselfattention,dasoulas2021lipschitznormalizationselfattentionlayers,yudin2025payattentionattentiondistribution} propose different strategies to enforce local Lipschitz continuity, which is a more relax condition.

While most prior analyses focus on convolutional or fully connected architectures, this work explores the direction of reinterpreting the Vision Transformer \cite{dosovitskiy2020image} (ViT) through dynamical systems theory. It opens the use of different theoretical tools from the dynamical systems theory, to construct a ViT model with a more robust, efficient, and explainable representations. In this work we further explore the \textit{autonomous setting} where the ode is time-invariant and only depend on the initial conditions, which is the vector representation of the input in this case. To the best of our knowledge, only~\cite{zhong_neural_2022} explored this autonomous setting experimentally, providing a natural baseline for comparison.

We propose \ode, a generalization of the attention block modeled as an ODE that encapsulates the necessary conditions to apply the different theoretical tools in the ViT leading a continuous-depth model that is both theoretically grounded and computationally efficient.
The main contributions of this paper are 
\begin{itemize}
    \item \textbf{ODE-ViT architecture:} We introduce a ViT-based formulation in which the attention block is reformulated as an ODE that satisfies the conditions for stability and well-posedness.  
    \item \textbf{Teacher–student training:} We propose a novel teacher–student framework in which a discrete ViT supervises the continuous trajectory of the ODE-ViT, aligning its intermediate states with the teacher’s representations.  
    This framework operates in a self-supervised manner and is specifically designed to replace the attention encoder of the ViT with our ODE-based formulation, making the proposed module fully plug-and-play.
    \item \textbf{Empirical validation:} Extensive experiments on CIFAR-10 and CIFAR-100 demonstrate that ODE-ViT achieves competitive performance with one order of magnitude fewer parameters, while offering improved interpretability and stability.
\end{itemize}

\section{Background}

As this work focuses on the ViT and the attention block;
let $\psi(x(t), t; \theta)$ be the attention block function parametrized by $t \in \mathbb{R}$ and the parameters $\theta$ 
 are the set all trainable parameter matrices $W$ where each $W_i$ lie in a $D$-dimensional space. The function $\psi(x(t), t; \theta)$ is split as \cite{zhong_neural_2022} with the Lie Trotter splitting scheme \cite{Blanes_Casas_Murua_2024} where the overall dynamics of the attention block is expressed as the composition of two sub-flows $F$ and $G$.
\begin{align}
    \psi(x(t), t, \theta) &= F(x(t),t, \theta) + G(x(t),t, \theta), \label{eq:split_ode}
\end{align}
In this paper $F(x(t),t;\theta)$ is the feedforward \texttt{MLP} and $G(x(t), t;\theta)$ is the {Attention block} function (\texttt{Attn}).
\begin{multline}
    \texttt{Attn}(X(t), t;\theta) = \texttt{sm}\!\left(
    \frac{XW_Q^{h} (W_K^{h})^{t} X^{t}}{\sqrt{d}}
    \right) XW_V^{(h)}
    \\ = \texttt{sm}(X A_h X^{t})\, XW_V^{h},
    \label{eq:attn}
\end{multline}
where $X \in \mathbb{R}^{N \times D}$ and $W_Q^{h}, W_K^{h}, W_V^{h} \in \mathbb{R}^{D \times d}$ are the projection matrices for head $h$,  
$\texttt{sm}$ denotes the row-wise softmax operation, and
\begin{equation}
    A_h = \frac{W_Q^{h} (W_K^{h})^{t}}{\sqrt{d}}.
    \label{eq:ah}
\end{equation}
Additionally, the \textbf{attention map matrix} $P_h$ for head $h$ is defined as:
\begin{equation}
    P_h = \text{sm}(X A_h X^{t}).
    \label{eq:ph}
\end{equation}

For proper notation of the ViT components, we define $\texttt{[CLS]}$ as the token class that is typically used in the ViT for the classification task. To guarantee the well-posedness of the ODE formulation and ensure stable trajectories, we analyze the Lipschitz continuity of $\psi(x(t), t; \theta)$.

A function $f : \mathbb{R}^D \rightarrow \mathbb{R}^d$ is said to be \textit{Lipschitz continuous} on a set $X \subseteq \mathbb{R}^D$ if there exists a constant $L > 0$ such that 
$\| f(x_1) - f(x_2) \|_2 \leq L \| x_1 - x_2 \|_2$ for all $x_1, x_2 \in X$.  
The smallest such constant $L$ is the \textit{Lipschitz constant} $\mathrm{Lip}(f; X)$.  
In practice, Transformers are not globally Lipschitz over $\mathbb{R}^D$ because of the unbounded norm of the Jacobian \cite{kim2021lipschitzconstantselfattention}, so we consider the weaker \textit{local Lipschitz continuity} within neighborhoods 
$\mathcal{B}_\varepsilon(x_0)$ around data points $x_0$.  
The local Lipschitz constant can be approximated by the spectral norm of the Jacobian, 
$\mathrm{Lip}(f; X) = \sup_{x \in X} \| J_f(x) \|_2$, ensuring local smoothness and bounded sensitivity to input perturbations.

As \cite{conf/nips/SanderAP22}, we define the distance between the discrete trajectory and the solution of associated ODEs as 
\begin{equation}\label{eq:min error}
    err = ||y(n) - x(n)||_2 ,
\end{equation}
\noindent
where $y$  is the solution of the associated ODE.

\begin{prop}[Approximation Error\cite{conf/nips/SanderAP22}]
\label{prop:approx_error}
Suppose that $\psi_\theta$ is $\mathcal{C}^1$ and $L$-Lipschitz with respect to $x$, uniformly in $t$. Let K be a compact space, $\psi_{t;\theta}(x(t)) = \psi(x(t), t; \theta)$ and let 
\[
C_N := \big\|\, \partial_t \psi_{t;\theta}(x(t)) + \partial_x \psi_{t;\theta}(x(t)) [\psi_{t;\theta}(x(t))] \,\big\|_{\infty,\, K \times [0,1]}.
\]
Then, for all $n$, the following approximation error bound holds:
\begin{equation}
    err \le
    \begin{cases}
        e^{L - 1}\, \dfrac{C_N}{2NL}, & \text{if } L > 0, \\[10pt]
        \dfrac{C_N}{2N}, & \text{if } L = 0.
    \end{cases}
    \label{eq:approx_error}
\end{equation}
\end{prop}
See that $\partial_t \psi_{t;\theta}(x(t)) + \partial_x \psi_{t;\theta}(x(t)) [\psi_{t;\theta}(x(t))]$ is the second order derivative of $\psi$, so for convenience we denote $\ddot{x}=\partial_t \psi_{t;\theta}(x(t)) + \psi_{t;\theta}(x(t)) [\psi_{t;\theta}(x(t))]$ and the upper bound approximation error can be written as $e^{L - 1}\, \dfrac{||\ddot{x}||_\infty}{2NL}$ where $||\ddot{x}||_\infty$ is the supremum of $\ddot{x}$.

\begin{obs}
This proposition implies that the solution of \eqref{eq:dynamic_equation} is well-defined, unique, and $\mathcal{C}^2$, and that its trajectory remains within some compact set $K \subset \mathbb{R}^d$.
\end{obs}

Finally, to further analyze the dynamical behavior and stability of the of the autonomous learned system, we compute the Lyapunov exponents, which quantify how small perturbations evolve along the learned flow.  
For a continuous system $\dot{x} = f(x)$, infinitesimal perturbations $\delta x(t)$ evolve according to $\delta \dot{x} = J_f(x(t))\,\delta x$, where $J_f$ is the Jacobian of $f$.  
The maximal Lyapunov exponent, $\lambda_{\text{max}}$, provides the dominant behavior, since all other exponents are necessarily smaller or equal.  
It is defined as:
\[
\lambda_{\text{max}} = \lim_{t \to \infty} \frac{1}{t} \log \frac{\|\delta x(t)\|}{\|\delta x(0)\|}.
\]
Where a negative maximal Lyapunov exponent indicates that all exponents are negative, meaning the system evolves toward a stable attracting invariant set.  
A positive value implies at least one exponentially diverging direction, typically associated with chaotic or unstable behavior.  
When $\lambda_{\text{max}} \approx 0$, the system neither expands nor contracts: if exactly one exponent is zero, the dynamics converge to a limit cycle.

From the Lyapunov exponent, we can define its inverse quantity $(\frac{1}{\lambda})$, the \textit{Lyapunov time}, which represents the characteristic time scale over which trajectories diverge. A smaller Lyapunov time indicates that instability arises more rapidly, while larger values correspond to smoother, more stable evolution in the latent manifold. 


\section{Methodology}
\label{sec:Methodology}
This section summarizes the \ode and how to force the \ode to follow the hidden representation of the original ViT. For the ViT, we used as a pretrained model DINO \cite{caron2021emergingpropertiesselfsupervisedvision} as its hidden representation is a generic representation not used for any specific task. We finetuned only the task-specific head for each dataset as we want to keep the generic representation. 

\subsection*{\ode}
\label{sec:odevit}

The ViT-Attention Encoder can be interpreted as a discretization of an ODE. In this work, we focus on adapting this continuous formulation to the Vision Transformer (ViT) architecture. The resulting model, denoted as \ode, is illustrated in \cref{fig:odeVit}. Given an input image $x \in \mathbb{R}^{H \times W \times C}$, the image is divided into patches of size $(S, S)$ using a convolutional layer, producing a sequence of $M+1$ patch embeddings, where $M = \frac{HW}{S^2}$ denotes the number of patches, and the additional ``$+1$'' corresponds to the \texttt{[CLS]} token.    

As previously discussed, the attention mechanism is not globally Lipschitz continuous~\cite{kim2021lipschitzconstantselfattention}.  
To address this, prior works~\cite{qi2023lipsformerintroducinglipschitzcontinuity,yudin2025payattentionattentiondistribution} have proposed various normalization strategies to control the spectral norm of the attention transformation, while ~\cite{kim2021lipschitzconstantselfattention} suggested replacing the standard dot-product attention with an $\ell_2$-normalized formulation to enforce bounded operator norms. Furthermore, a recent work~\cite{yudin2025payattentionattentiondistribution} force a local Lipschitz condition by regularizing $P_h$ with a loss function.
\begin{equation}\label{eq:JaSMin}
    \mathcal{L}_{JaSMin_k} = \sum^{L, H}_{l, h} \text{max log}\left( \frac{g_1(P^{l, h}_{i,:})}{g_k(P^{l, h}_{i,:})} \right), \; k>1,
\end{equation}
\noindent
where $g_k(P_{i,:})$ is the $k$ largest component of $P_{i}$ for each layer and  head.  This loss penalizes deviations in the eigenvalue distribution of the attention matrix, effectively constraining its spectral radius and promoting stable, Lipschitz-consistent behavior across heads and layers.

Our architecture condenses all these mechanisms to generalize the attention block as an ODE while maintaining the dot-product attention. Following~\cite{qi2023lipsformerintroducinglipschitzcontinuity}, all projection weights are initialized based on their maximum singular value $||W||_2$, thereby bounding the spectral norm of the linear transformations. Furthermore, standard normalization layers (e.g., LayerNorm) are replaced with \textbf{center normalization} to stabilize the variance of intermediate representations.

The parameters $\theta$ remain shared across all iterations to force the autonomous setting. Since $\psi$ is designed to be Lipschitz continuous, the existence and uniqueness of the solution are guaranteed as $N \to \infty$, following the Picard–Lindelöf theorem.

\subsection*{Teacher Representation as a Solution of the Differential Equation}

\begin{figure}[t]
    \centering
    \includegraphics[width=\columnwidth]{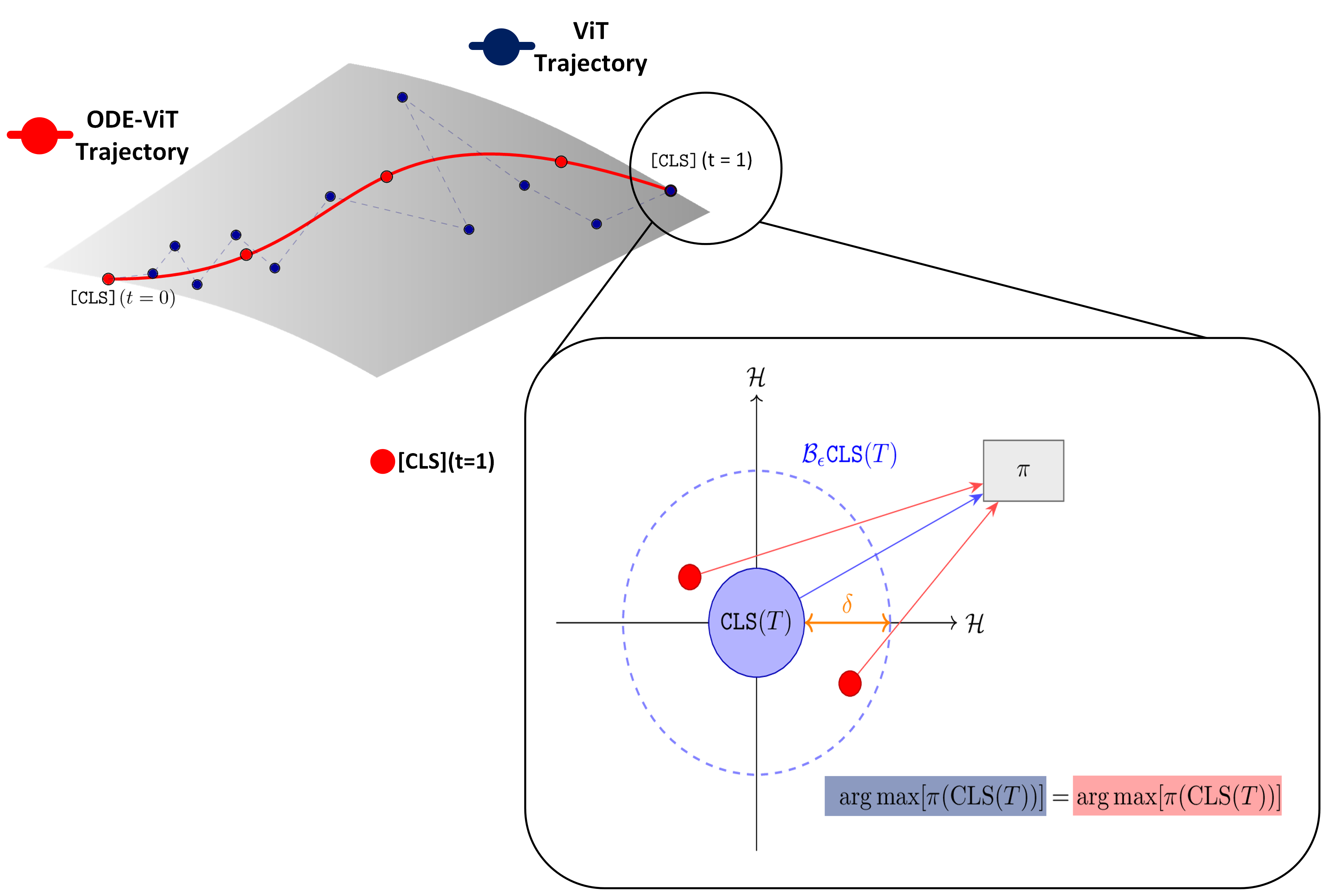}
    \vspace{0.5em}
    \caption{Exemplification of the hypothesis that the MSE does not need to converge to the same point as the teacher, but only within the contraction region. In this region, the classification head predicts the correct class identically to the teacher. Both CLS(T) refer to the last representation of both attention blocks.}
    \label{fig:invariance}
\end{figure}
\begin{figure*}[t]
    \centering
    \includegraphics[width=1.05\linewidth]{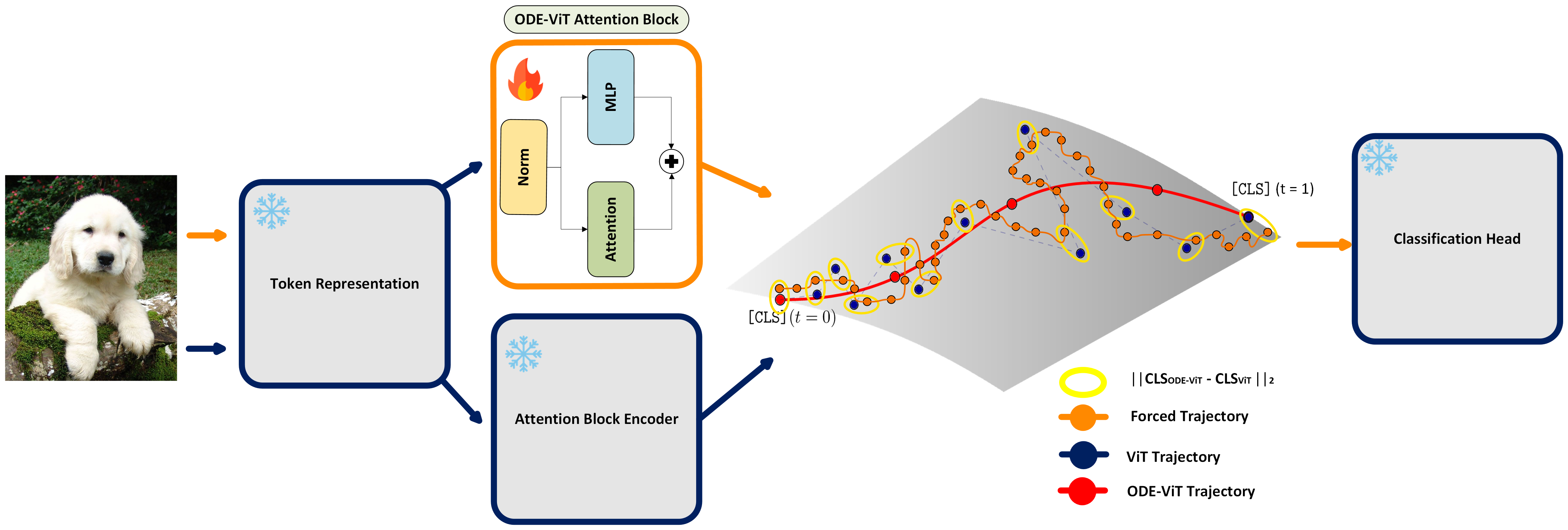}
    \caption{
            Overview of the \textbf{plug-and-p lay Teacher--Student framework}.  
            The input image is processed through both Token Representation modules and passed through their respective attention layers.  
            The orange trajectory represents the \ode architecture with teacher supervision, while the blue pathway corresponds to the Teacher ViT. The forced trajectory is optimized by minimizing \eqref{eq:min error} from the \texttt{[CLS]} tokens. The classification head ($\pi$) is initialized from the teacher and kept frozen during training; in our experiments, only one case benefited from unfreezing it. It is also exemplified that the proportion of states required to reach each ViT hidden state varies across layers.
            }
    \label{fig:Teacher-Student Framework}
\end{figure*}

We do not know which is the trajectory that reaches the best solution, however we know that the last representation of our pretrained ViT is a specific solution along a trajectory. With that in mind, we force our \ode to learn which is the trajectory that passes through the last hidden representation of the pretrained ViT. In fact, by \cite{conf/nips/SanderAP22}, we know that we can learn that trajectory bounded by \eqref{eq:approx_error}.
Under this assumption, we defined a ViT teacher model and used \ode as the student to match the teacher last hidden representation to the \ode last state by minimizing a combination of Mean Squared Error (MSE) and JaSMin \eqref{eq:JaSMin} losses. We work under the assumption that the last state of the teacher ViT hidden representation lies within a contraction region, so any local perturbation $\delta$ around this point satisfies that 
\begin{equation}\label{eq: invariance classification}
	\pi(\psi(x, N, \theta) \approx \pi(\phi(x, T, \theta)
\end{equation}
where $\pi$ denotes the projection onto the classification head of the  head and $\phi$ is the attention block from the teacher ViT. In other words, all neighboring representations in the contraction region are mapped to approximately the same output of the teacher's classification projection; thus, the result of classifying that point should be the same. The intuition can be seen in \cref{fig:invariance} and the Teacher-Student framework is depicted in \cref{fig:Teacher-Student Framework}

\section{Experiments}
\label{sec:Experiments}
To experimentally validate the working hypothesis, we conducted different experiments in different scenarios and datasets. 
We re-evaluated the experiments carried out in \cite{zhong_neural_2022} to reproduce the autonomous setting results.

\subsection{\ode}
In \cite{ott2021resnet}, authors discuss whether a properly trained ODE is invariant to variations in the number of steps, noting that such changes may lead to a slight decrease in performance, but are robust overall. In that work, the complexity of the problems was controlled, and the results were mostly expected. We further validated the experimental set up to see if a wider and more complex dataset converge to different results or, for instance, can be interpreted in the same way as \cite{ott2021resnet}. In our case, we confirm it by evaluating the trained model with larger and smaller $N$. The figure is depicted in the supplementary material. 

\begin{table*}[]
\centering
\resizebox{0.8\textwidth}{!}{%
\begin{tabular}{@{}ccccccc@{}}
\toprule
                                                         & Dataset                         & Param (M) & Losses        & Acc@1 & Acc@3 & Acc@5 \\ \midrule
\multirow{3}{*}{\begin{tabular}[c]{@{}c@{}}Teacher\\ DINO-base + Training Only Head\end{tabular}} & Cifar10 & 85 & CE & 0.923 & 0.993 & 0.997 \\
                                                        & Cifar100                        & 85        & CE            & 0.881 & 0.968 & 0.982 \\
                                                        & Imagenet100                     & 85        & CE            & 0.923 & 0.981 & 0.99  \\ \midrule
\multirow{1}{*}{ODE \cite{zhong_neural_2022}}           & Cifar100                        & 0.7       & CE            & 0.533 & -     & -   \\ \midrule
\multirow{3}{*}{\ode}                                   & Cifar10                         & 0.5       & CE            & 0.809 & 0.98  & 0.99  \\
                                                        & Cifar100                        & 4.2       & CE            & 0.579 & 0.728 & 0.794 \\
                                                        & \multicolumn{1}{l}{Imagenet100} & 7         & CE            & 0.513 & 0.701 & 0.754 \\ \midrule
                                                        & Cifar10                         & 7         & MSE+JasMin    & 0.885 & 0.980 & 0.992 \\
\ode Teacher-Student Base                               & Cifar100                        & 7         & MSE+JasMin+CE & 0.721 & 0.872 & 0.914 \\
                                                        & \multicolumn{1}{l}{Imagenet100} & 7         & MSE+JasMin    & 0.684 & 0.817 & 0.865 \\ \midrule
\multirow{2}{*}{\ode Teacher-Student Small}             & Cifar10                         & 3.8       & MSE+JasMin    & 0.867 & 0.973 & 0.991 \\
                                                        & Cifar100                        & 3.8       & MSE+JasMin    & 0.657 & 0.819 & 0.914 \\ \bottomrule
\end{tabular}%
}
\caption{Best results obtained by the \ode model with and without the proposed teacher--student framework.  
The teacher corresponds to the DINO-Base model, where only the classification head is trained.  
The listed losses correspond to the training signals used to optimize the ODE. The difference in the \ode base and small is in the MLP, reducing the up-projection in the intermediate layer.
}
\label{tab:Teacher-Student Results}
\end{table*}

Overall, there is a distance in the results of the \ode ViT. However, our ODE-ViT training has not been trained with transfer learning but from scratch. We notice that, in fact, the results of a ViT with the same number of parameters as our \ode trained without transfer learning show equivalent results. The results are in \cref{tab:Less Params results}, where it can be seen that the distance in the results is negligible when the ViT has similar size as \ode.

\subsection*{Teacher Representation as a Solution of the ODE}

As discussed in Section~\ref{sec:Methodology}, we investigate whether the performance gap between a classical ViT and its ODE-based counterpart arises from the fact that the underlying ODE solution is not explicitly known.  
Suppose the attention block of a standard ViT can indeed be interpreted as the discretization of a continuous dynamical system. In that case, the final hidden representation of the ViT should lie on (or close to) the trajectory generated by the corresponding ODE. For instance, the \ode should be capable of passing through,  or near this representation under \eqref{eq:approx_error}.  

This approximation error quantifies the convergence of the proper loss. We notice that when the MSE from the trajectory to the hidden representation of the pretrained model is below \eqref{eq:approx_error}, the loss function starts to be in a plateau.

Following  ~\eqref{eq: invariance classification}, if the \ode reaches a neighborhood of the teacher’s final hidden state, the same classification head can be reliably used for prediction.

To validate this hypothesis, we initialized all parameters of the \ode with those of the pretrained teacher ViT, except for the attention block, which is replaced by the ODE-based attention module.  The input image is passed through the same patch embedder, and the output token sequence is feed to the attention block encoder of both \ode and teacher. Finally, the last state of the \ode trajectory is the input of the classification head. During training, all parameters were frozen except for those within the ODE-attention block.

Matching the hidden representations between the teacher and student models is non-trivial, since the latent states of the teacher evolve under different parameterizations and occupy distinct regions of the representation manifold.  
To address this matching, we first computed the mean pairwise distance between consecutive hidden representations in the teacher ViT, and used this distances to define a set of \textit{checkpoints} along the trajectory.  
Specifically, we applied a softmax normalization over the mean pairwise distance between consecutive layers to estimate the relative progression that the trajectory need to reach each of the hidden representations starting from the same starting point ($\texttt{[CLS]}_{\ell=0}$) as the teacher in the representation space.

\begin{figure*}[t]
    \centering
    \includegraphics[width=\linewidth]{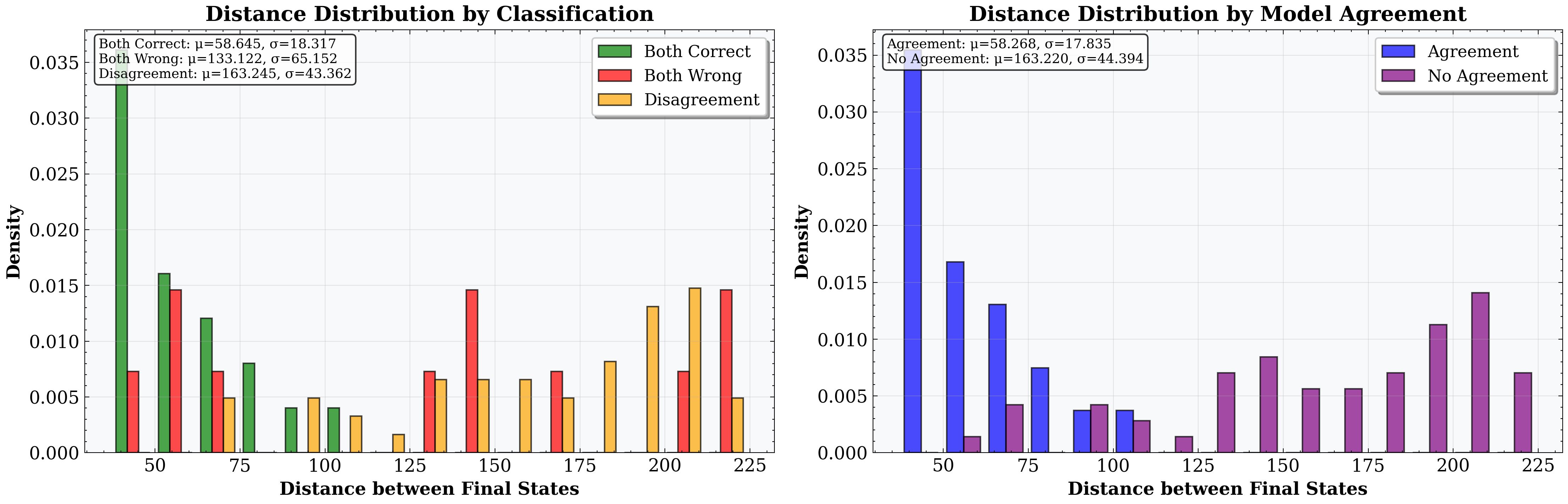}
    \caption{Analysis plot with the distances between the \texttt{[CLS]} token of the ViT and the \texttt{[CLS]} token of the \ode for all samples in the test set. The green color means that the \ode and the ViT correctly classify the sample, while the red means that both models badly classify it. Finally, the yellow means that only the teacher correctly classify the sample.}
    \label{fig: contraction results}
\end{figure*}

The results of these experiments, conducted across multiple datasets and input signals, are reported in \cref{tab:Teacher-Student Results}.  
Empirically, we observe that the teacher representations, particularly \( \texttt{[CLS]}_{te}^{\ell-1} \), lie within a contractive region of the learned dynamical system. This finding is illustrated in \cref{fig: contraction results}, where we show that \texttt{[CLS]} tokens that are spatially close in the latent manifold are assigned the same class by the student model, even though the training was performed without supervision, solely using the MSE and JaSMin losses.

We worked under the assumption that if \ode has truly learned the internal dynamics of the system, it should be robust to variations in the initial conditions. To test this hypothesis, we computed the Lyapunov exponents for the different models to examine whether larger Lyapunov exponents correlate with misclassified samples.  
The results can be seen in \cref{fig:Lyapunov_Distribution}, where they show exponents relatively low, suggesting that the model is robust with respect to small perturbations in the initial conditions (IVPs).  
Furthermore, we observe a slight correlation between classification accuracy and the Lyapunov exponent: classes with lower accuracy tend to exhibit slightly higher Lyapunov exponents. In fact, by computing the inverse of the Lyapunov exponent, we can quantify the specific step at which the attention block begins to exhibit signs of instability.  

This behavior suggests that when the dynamics of the teacher model are locally contractive, the resulting feature representation is robust to small perturbations in the hidden space. In practice, this implies that once the model has reached a stable attractor in the representation manifold, the classification head becomes robust to minor deviations in the latent representation. It means that we can train the \ode only to mimic the \texttt{[CLS]} representation of the ViT with no additional losses than the MSE and the JasMin \citep{yudin2025payattentionattentiondistribution} in a self-supervised way.
The results of this hypothesis can be seen in \cref{fig: contraction results}. There is a clear correlation in the distance between both \texttt{[CLS]} tokens, where up to a certain distance the frozen classification head from the teacher classifies the same both \texttt{[CLS]} tokens from \ode and the teacher model. 
Finally, we add the comparison from our \ode to similar models with a similar number of parameters as our model in order to compare the distance to them. The results can be seen in \cref{tab:Comparison Results} where it is shown that our model performs in similar to other ViT models with similar sizes.

\begin{table}[t]
\centering
\resizebox{\columnwidth}{!}{%
\begin{tabular}{@{}lccccc@{}}
\toprule
                            & Dataset  & \multicolumn{1}{l}{Params (M)} & Acc@1 & Acc@3 & Acc@5 \\ \midrule
ViT from scratch            & Cifar10  & 6.8                       & 0.909 & 0.99  & 0.99  \\
                            & Cifar100 & 6.8                       & 0.665 & 0.84  & 0.87  \\ \midrule
\multicolumn{1}{c}{\ode} & Cifar10  & 0.5                       & 0.809 & 0.98  & 0.99  \\
                            & Cifar100 & 4.2                       & 0.579 & 0.72 & 0.79 \\ \bottomrule
\end{tabular}%
}
\caption{Comparison between a ViT trained from scratch with similar size as our trained \ode from scratch. The \ode competes to ViT when the models are trained without transfer learning from pretrained weights.}
\label{tab:Less Params results}
\end{table}

\begin{table}[t]
\centering
\resizebox{\columnwidth}{!}{%
\begin{tabular}{@{}cccccc@{}}
\toprule
                                                        & Param (M) & Losses     & Acc@1        & Acc@3 & Acc@5 \\ \midrule

HSViT-C3A4 \cite{xu2024hsvithorizontallyscalablevision} & 3         & CE         & 0.738  & -     & -     \\
MSCVIT-T \cite{zhang2025mscvitsmallsizevitarchitecture} & 4         & CE         & 0.8 & -     & -     \\ \midrule
DeiT Tiny \cite{pmlr-v139-touvron21a}                   & 5.72      & CE + KL        & 0.644        & -     & 0.892 \\ 
AttentionProbe \cite{wang2022attention}                 & 2.31      & Probe + CE + KL & 0.716 & -    & - \\
\ode Teacher-Student                                    & 7         & MSE+JasMin & 0.721  & 0.872 & 0.914 \\ \bottomrule
\end{tabular}%
}
\caption{Comparison of our \ode trained with the Teacher--Student framework and ViTs of similar sizes over the CIFAR-100. The two first models are variations of the proper architecture of the ViT while the other ones are trained with a knwoledge distillation strategy. }
\label{tab:Comparison Results}
\end{table}

\section{Related Work \& Discussion}  
The interpretation of residual networks as discretized ordinary differential equations (ODEs) was first formalized in~\cite{chen2019neuralordinarydifferentialequations}, demonstrating that residual connections can be viewed as Euler approximations of continuous flows. Other works extended this perspective by analyzing the stability and expressivity of continuous-depth models~\cite{li2020deeplearningdynamicalsystems,ott2021resnet,teshima2020universalapproximationpropertyneural,conf/nips/SanderAP22}. Most of these previous analyses focuses on convolutional and fully connected architectures \cite{Queiruga2020ContinuousinDepthNN} or ViT in a non-autonomous setting where the dynamics of the system are time-dependent \cite{tong_neural_2025, zhong_neural_2022} and the parameters that model the continuous trajectory are different. 

The non-autonomous flow adds additional degrees of freedom and makes it harder to apply and interpret the different dynamical systems tools. Furthermore, different parameters imply that the trajectory is build from different functions, which, without a theoretical proof, is not trivial. In ResNets, each encoding block belongs to a non-autonomous setting, since the representation space changes across layers due to bottlenecks and down-sampling operations, and it is naturally convenient to model this architecture in that way.

However, Transformers maintain a constant representation space across layers; thus, introducing additional parameters per block directly affects the properties of the ODE and adds further complexity that is not strictly needed in our formulation.
To the best of our knowledge, only~\cite{zhong_neural_2022} has experimentally explored the autonomous formulation, showing results for a single independent attention layer within the ViT framework.  However, they only do the experiment without additional details of their experimental set up, methodology, and discussion. They focus only on the conceptual connection between attention layers and ODEs without taking into consideration the needed conditions to make these modules fit the theory. 

\cref{tab:Less Params results} shows the results of \ode training from scratch and response a natural question; Does \ode performs similarly good as a ViT with a similar size? 
In \cref{tab:Less Params results} is also shown the comparison between \ode and a ViT trained in the same conditions to validate that \ode can scale, and indeed it does.

\ode is constructed upon the most fundamental formulations from the dynamical systems perspective, employing the Euler method and the Lie–Trotter splitting scheme.
Other works, such as~\cite{lu*lu_understanding_2019}, have explored more advanced schemes such as the Strang splitting method~\cite{Blanes_Casas_Murua_2024}.  
However, we consider the Strang splitting scheme to represent a natural next step for future exploration within this research line.
The theoretical framework and tools employed in this work rely on the assumption that the attention block within the encoder is Lipschitz continuous.  
Although this assumption has been shown not to hold in general~\cite{kim2021lipschitzconstantselfattention}, recent studies have proposed methods to partially address this issue, demonstrating that the attention block can be made locally Lipschitz~\cite{qi2023lipsformerintroducinglipschitzcontinuity,castin2024smoothattention,yudin2025payattentionattentiondistribution}.  
Our formulation incorporates these insights by constraining the spectral radius of the attention weights and replacing standard normalization with centered normalization layers, which results in smoother and more stable trajectories in the latent manifold.  
In practice, these modifications significantly improve the stability of MSE-based training, preventing representation collapse in intermediate states.

\begin{figure*}[t]
    \centering
    \begin{minipage}[t]{0.48\linewidth}
        \centering
        \includegraphics[width=\linewidth]{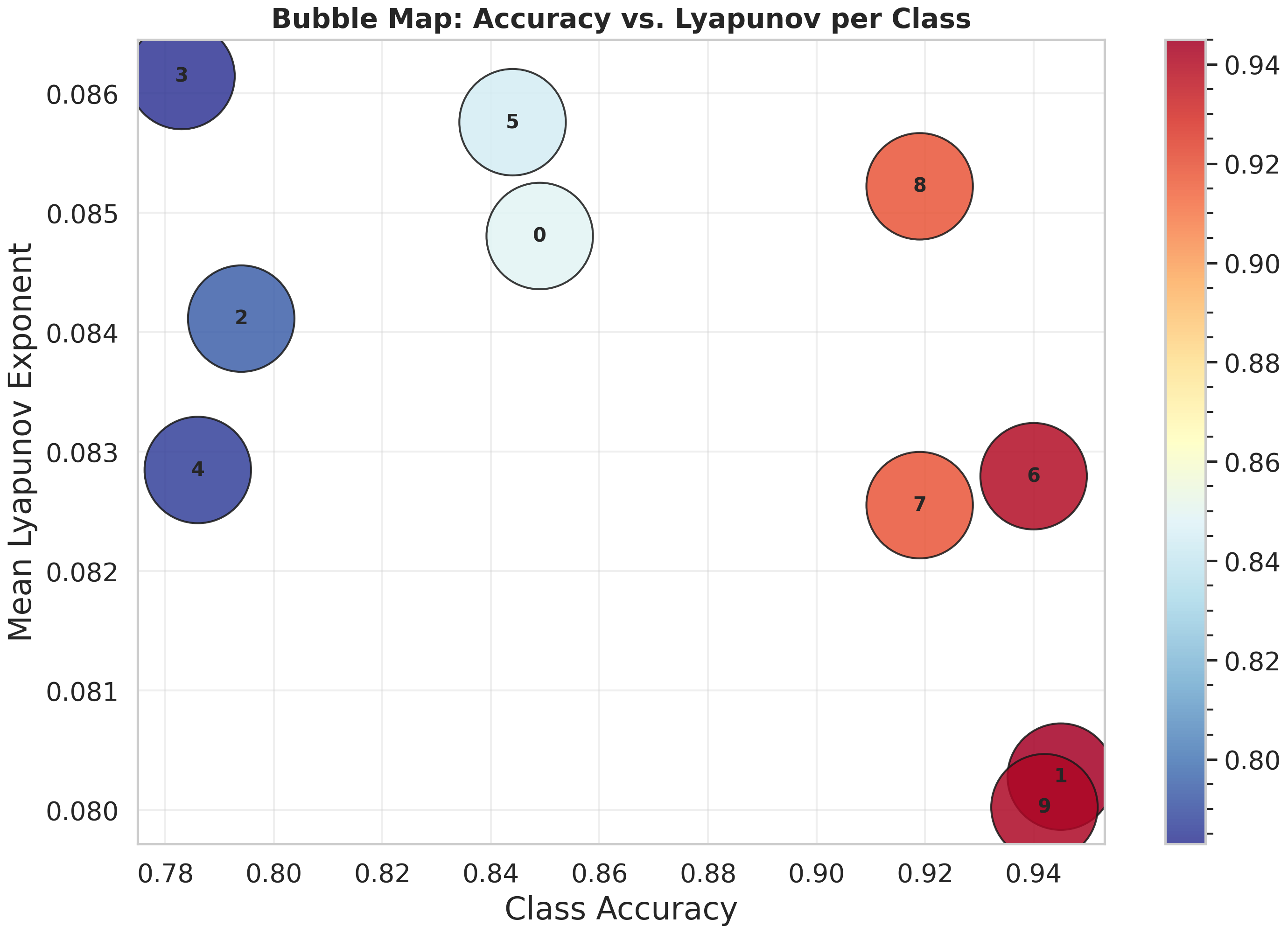}
        \label{fig:Invariance CIF10}
    \end{minipage}
    \hfill
    \begin{minipage}[t]{0.48\linewidth}
        \centering
        \includegraphics[width=\linewidth]{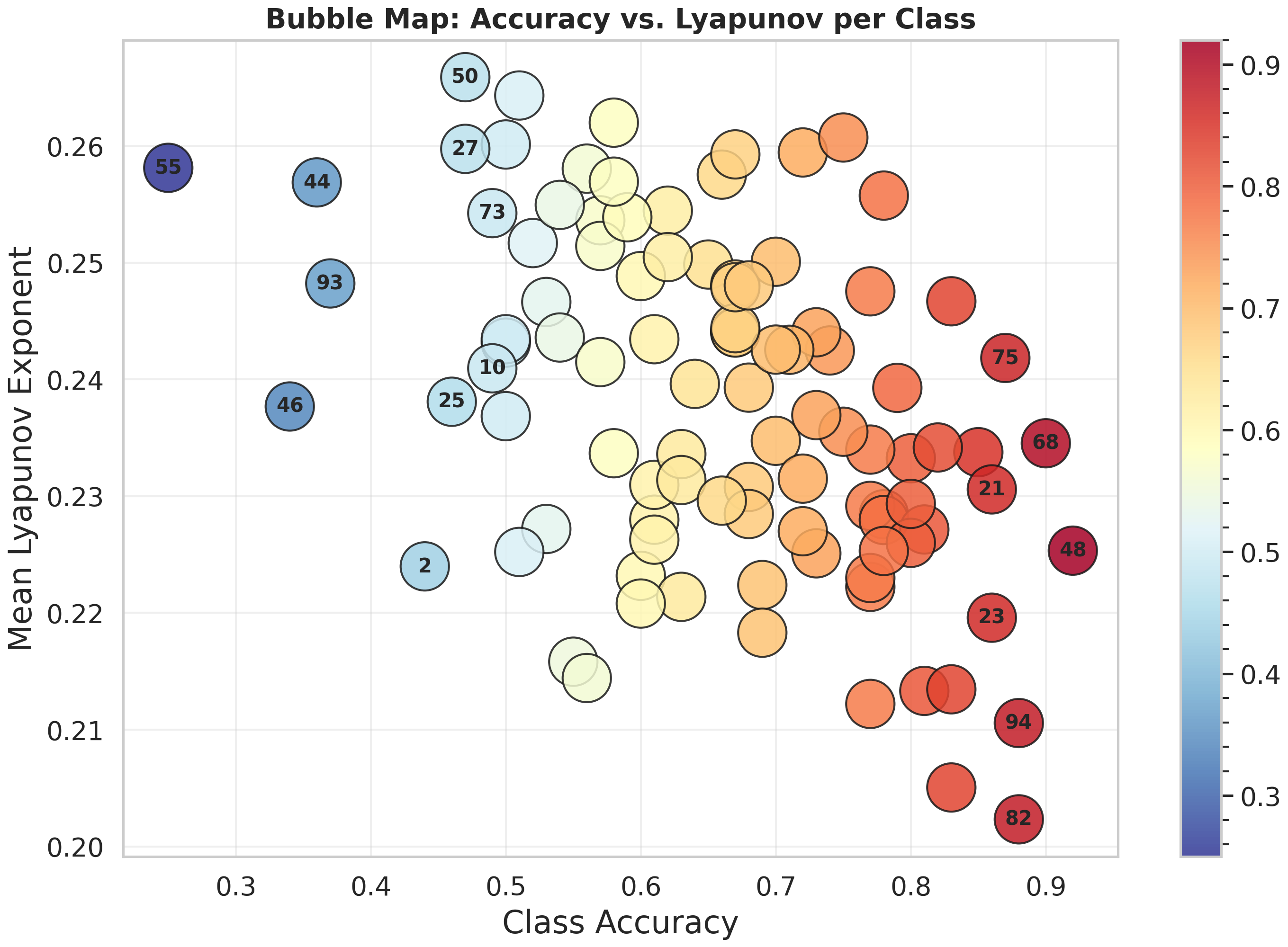}
        \label{fig:Invariance CIF100}
    \end{minipage}

    \caption{
    Distribution of Lyapunov exponents for \textbf{CIFAR-10} (left) and \textbf{CIFAR-100} (right) classes.  
    The color and horizontal position of each bubble represent the classification accuracy of the ODE-ViT for the corresponding class, while the vertical axis shows the mean Lyapunov exponent of its dynamics.  
    In both datasets, a mild correlation emerges between the Lyapunov exponent and the classification accuracy, indicating that classes with more stable dynamics (lower Lyapunov exponents) tend to exhibit higher accuracy. 
    }
    \label{fig:Lyapunov_Distribution}
\end{figure*}

Our experiments with our formulation provide both theoretical and empirical evidence supporting the continuous-time interpretation of the Vision Transformer. In particular, this theoretical foundation enables the use of a teacher–student framework, where the \ode aligns its continuous trajectory with the latent dynamics of a discrete ViT.  
As shown in \cref{tab:Teacher-Student Results}, the intermediate representations of the teacher effectively serve as checkpoints along the ODE trajectory, guiding the student model toward stable and consistent dynamics. This training strategy boost the performance 8\% in CIFAR-10, 14\% in CIFAR-100 and 16\% in Imagenet-100. Although the results are still far from the teacher model, more complex attention functions or integrators can be used to improve the results, as \ode is build as a baseline.

The teacher-student training also demonstrates that training can be performed in a self-supervised manner: since the classification head lies within a contraction region, only the attention block requires replacement and training, making our approach effectively \textit{plug-and-play}. Within this teacher–student framework, we further observed that model stability can be quantified by tracking the upper bound of the approximation error in \eqref{eq:approx_error}.  
Each of the MSE losses begins to converge when the MSE between \texttt{[CLS]} token from the \ode and the teacher of every checkpoint is below \eqref{eq:approx_error}, which is something firstly formulated in \cite{conf/nips/SanderAP22}. 

We studied the stability of the system using the Lyapunov exponents, as shown in \cref{fig:Lyapunov_Distribution} and discussed in~\cite{tong_neural_2025}. Classes with smaller accuracy are slightly correlated with more unstable trajectories within the representation space. The Lyapunov exponents provide a quantitative measure of the onset of instability in the model’s dynamics. By computing the \textit{Lyapunov time}, we could identify the time step at which the trajectory begins to diverge.  This allows us to determine the precise point where the system transitions from stability to instability and, consequently, when the evaluation of the function $\psi$ can be safely terminated, avoiding unnecessary computation. 
  
\begin{figure}[t]
    \centering
    \includegraphics[width=\columnwidth]{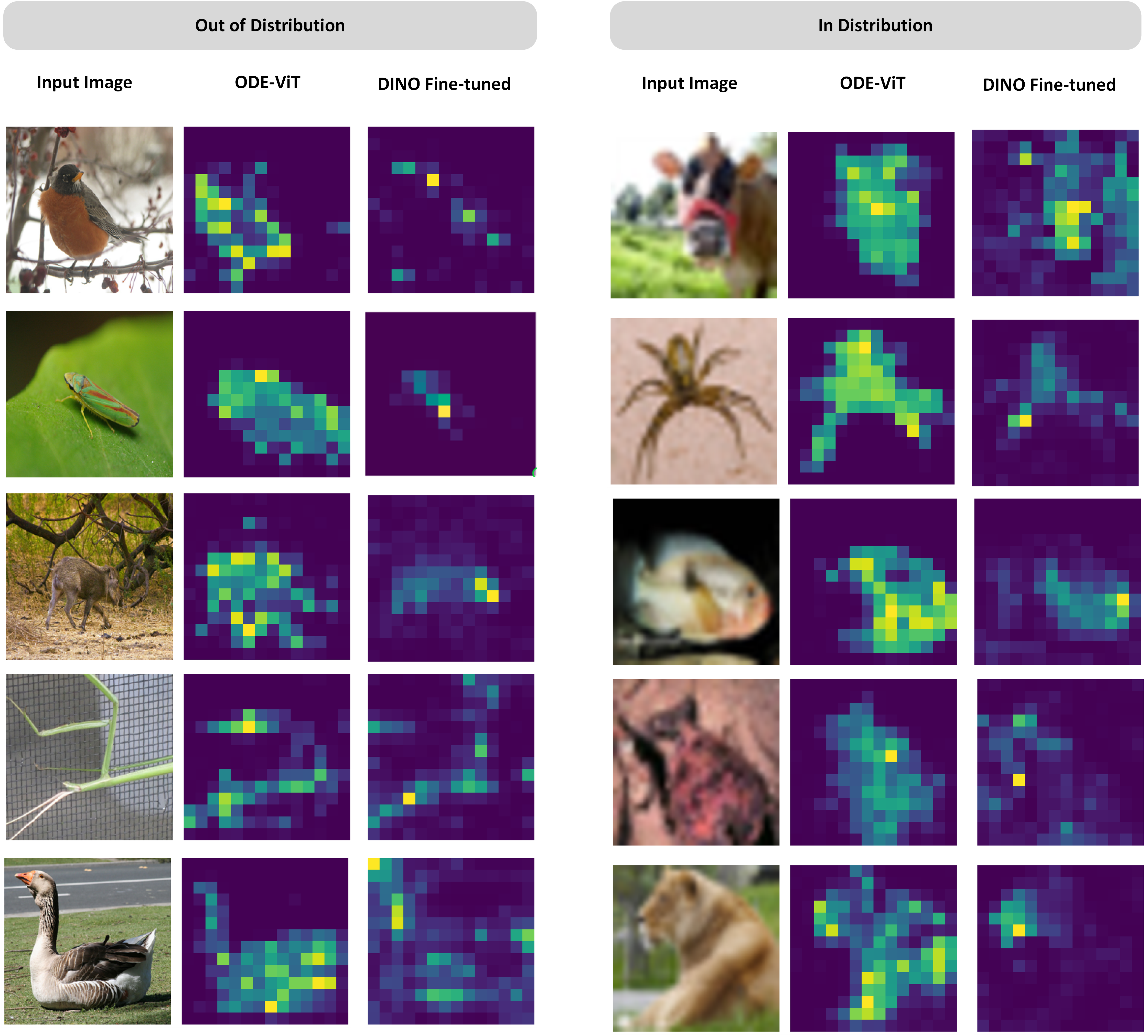}
    \caption{Attention maps of the \texttt{[CLS]} token from the \ode at its final state. Images in the left are taken from the ImageNet dataset, while the images in the right are taken from CIFAR-100. The model used to extract these representation was trained on CIFAR-100 using the \ode Base after trained with the teacher-student framework. The results demonstrate that the model has learned a generalizable representation.}
    \label{fig: Attentions}
\end{figure}

The hypothesis about the stability and the contraction region are supported by the results in \cref{fig: contraction results} and \cref{tab:Teacher-Student Results}, where models with smaller distances between checkpoints achieve strong classification performance without requiring the cross-entropy signal for supervision. Furthermore these models maintain or even improve the attention distribution as the DINO-teacher model \cref{fig: Attentions}.

The reduction of the parameters and the computation is a behavior that appears naturally in this work, as our aim is not to reduce as maximum possible the parameters, but generate an efficient, robust, and interpretable architecture from the theoretical point of view. In \cref{tab:Comparison Results} we have shown a comparison of ViT models with a similar number of parameters as \ode. The models that have modified the attention block architecture achieve the best performance. In the future we consider adapting this formulations in our \ode function $\psi$ to confirm that they are suitable to model an ODE. In the other hand, we perform similar to other models with similar parameters that have been trained with distillation strategies, however our framework can be trained in a self-supervised way and it simpler to adapt.
  
\section{Conclusions}
\label{sec:conclusion}

In this work, we introduced \textbf{\ode}, a principled reformulation of the Vision Transformer (ViT) under the lens of ordinary differential equations (ODEs).  
By enforcing local \textit{Lipschitz condition} and ensuring well-posed autonomous dynamics, we show that the Transformer attention mechanism can be effectively modeled as a continuous-time dynamical system.  
This theoretical grounding enables a rigorous analysis of model stability and convergence, bridging discrete ViTs with continuous-depth formulations.
Our experiments confirm that the proposed formulation maintains stability and achieve competitive performance with significantly fewer parameters, which without being the main focus of the approach, happens naturally.  
Moreover, the proposed \textbf{teacher-student framework} effectively transfers discrete ViT dynamics into the continuous domain in a self-supervised set up, narrowing the performance gap while preserving interpretability and theoretical soundness. Beyond empirical improvements, this study establishes a experimental foundation to build on top.
In future work, we plan to explore higher-order numerical solvers, adaptive step-size strategies, and extensions to multi-modal architectures, further leveraging continuous formulations to advance the understanding and performance of modern Transformer models, thereby building more efficient and transparent architectures.

{
    \small
    \bibliographystyle{ieeenat_fullname}
    \bibliography{main}
}


\clearpage
\setcounter{page}{1}

\twocolumn[{%
\renewcommand\twocolumn[1][]{#1}%
\maketitlesupplementary
\centering
\captionsetup{type=figure}
    \begin{tabular}{ll}
    \includegraphics[width=\columnwidth]{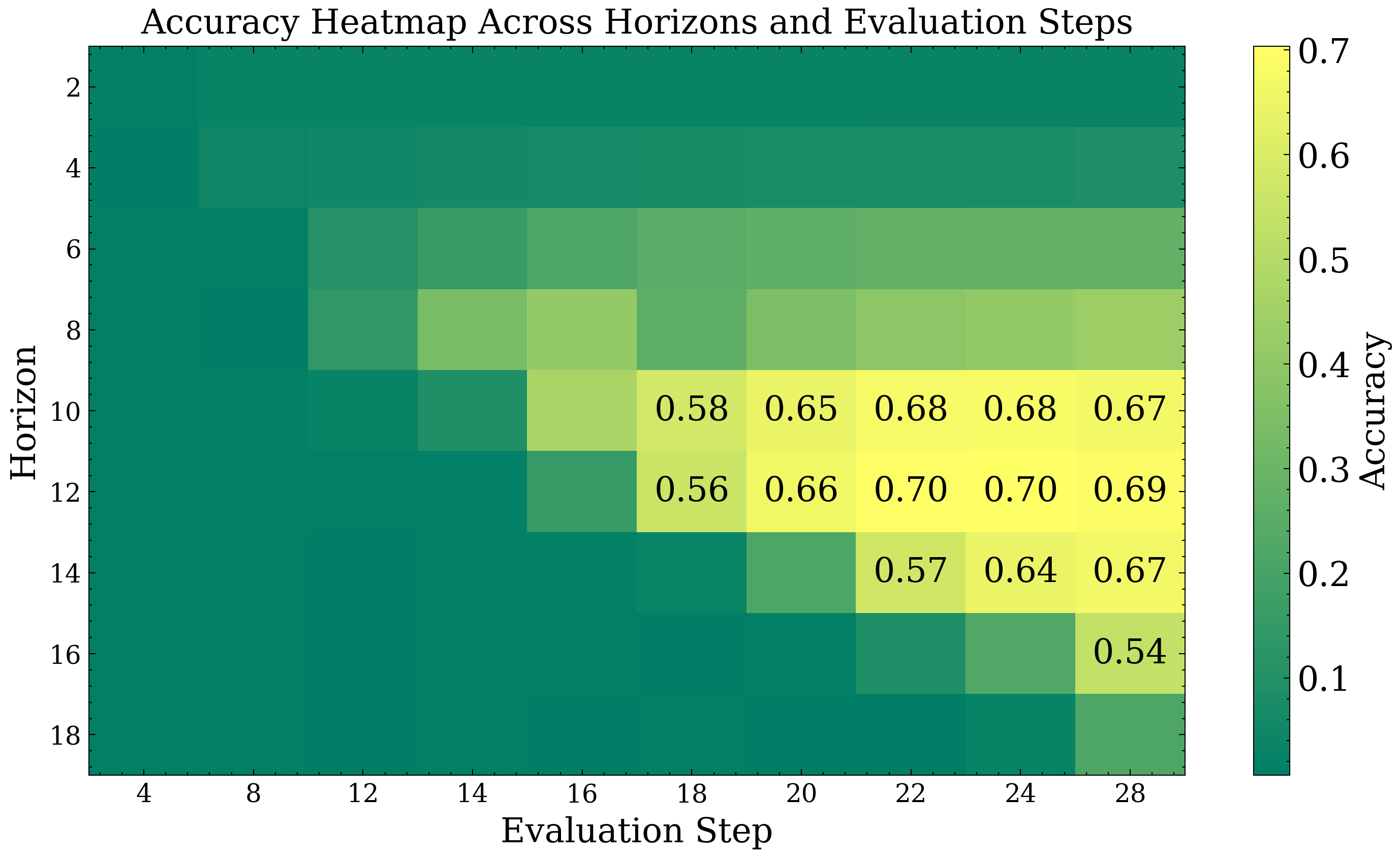}
    &
    \includegraphics[width=\columnwidth]{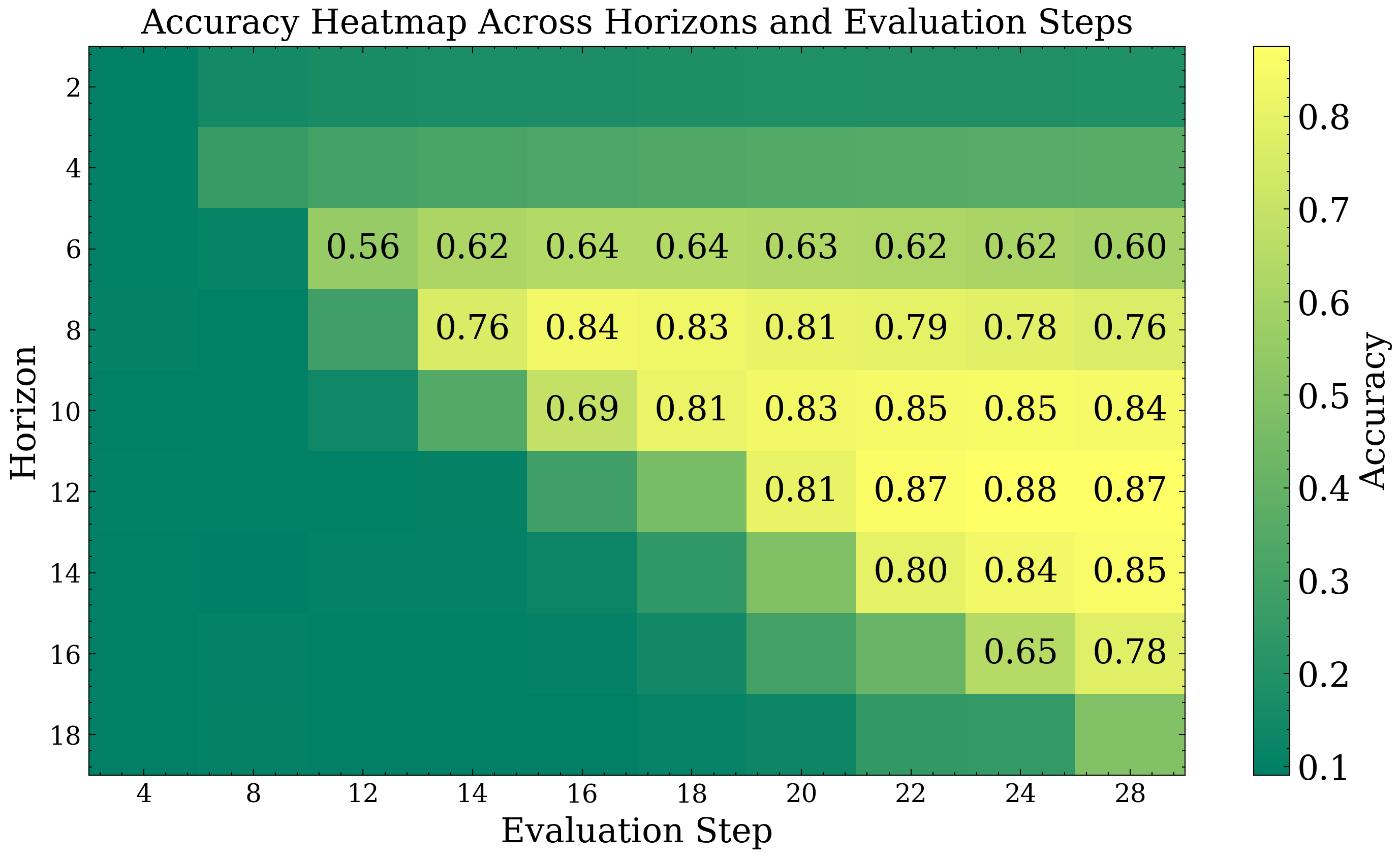}
    \end{tabular}
    \caption{The heatmaps show the correlation among different resolutions and horizons with the accuracy on, from left to right, CIFAR-100 and CIFAR-10. It can be seen that with less resolution in a lower horizon, the results are similar in both cases. The horizon number is related to the profundity of the ViT, where 12 is the one used to train to emulate the ViT depth.}
    \label{fig: horizon evaluations}
\vspace{5mm}
 
}]

\begin{table*}[t!]
\centering
\resizebox{\linewidth}{!}{%
\begin{tabular}{@{}ccccccccccccccc@{}}
\toprule
 &
  Dataset &
  Param (M) &
  Losses &
  Acc@1 &
  Acc@3 &
  Acc@5 &
  Image Size &
  Patch Size &
  Embedding Size &
  MLP ratio &
  Eval Steps &
  Register Tokens &
  pos\_embed\_register\_tokens &
  Heads \\ \midrule
\multirow{3}{*}{\ode} &
  Cifar10 & 0.5 & CE & 0.809 & 0.98 & 0.99 & 224 & 16 & 192 & 1 & 24 & 1 & No & 8 \\
 & Cifar100    & 4.2 & CE         & 0.579 & 0.728 & 0.794 & 224 & 16 & 768 & 1 & 24 & 1  & No & 8  \\
 & Imagenet100 & 7   & CE         & 0.513 & 0.701 & 0.754 & 224 & 16 & 768 & 1 & 24 & 1  & No & 12 \\ \midrule
 & Cifar10     & 7   & MSE+JasMin & 0.885 & 0.980 & 0.992 & 224 & 16 & 768 & 4 & 24 & 10 & No & 12 \\
\ode Teacher-Student Base &
  Cifar100 &   7 &   MSE+JasMin+CE &   0.721 &  0.872 &  0.914 &  224 & 16 & 768 & 4 & 24 & 10 & No &  12 \\
 & Imagenet100 & 7   & MSE+JasMin & 0.684 & 0.817 & 0.865 & 224 & 16 & 768 & 4 & 24 & 10 & No & 12 \\ \midrule
\multirow{2}{*}{\ode Teacher-Student Small} &
  Cifar10 &   3.8 &  MSE+JasMin &  0.867 &  0.973 &  0.991 &  224 &  16 &  768 &  1 & 24 &  10 &  No &  12 \\
 & Cifar100    & 3.8 & MSE+JasMin & 0.629 & 0.792 & 0.914 & 224 & 16 & 768 & 1 & 24 & 10 & No & 12 \\ \midrule 
\end{tabular}%
}
\caption{Table with the hyperparameters used to extract these performances. In the teacher-student framework, the gap of the parameters arises from the frozen parameters of the classification head and the patch token extraction module.}
\label{tab: Hyperparameters}
\end{table*}

This supplementary material provides additional information about the experimental setup and extra experiments that address claims made in the paper.

\section{Experimental Set up}
\label{sec:Experimental Set up}

All experiments were conducted on an NVIDIA A40 GPU with 45\, GiB of memory. We used a batch size of 64, and optimization was performed using the AdamW optimizer with an initial learning rate of \(1\times 10^{-4}\) and a weight decay of \(5\times 10^{-2}\). The learning rate scheduler was \textit{get\_cosine\_with\_hard\_restarts\_schedule\_with\_warmup}  from the HuggingFace framework, configured with a warmup phase corresponding to 
10\% of the total training epochs and 10 hard restart cycles. The total epochs done to train the different models are 300.

\begin{figure*}[t]
\begin{tabular}{ll}
\centering
\includegraphics[width=0.95\columnwidth]{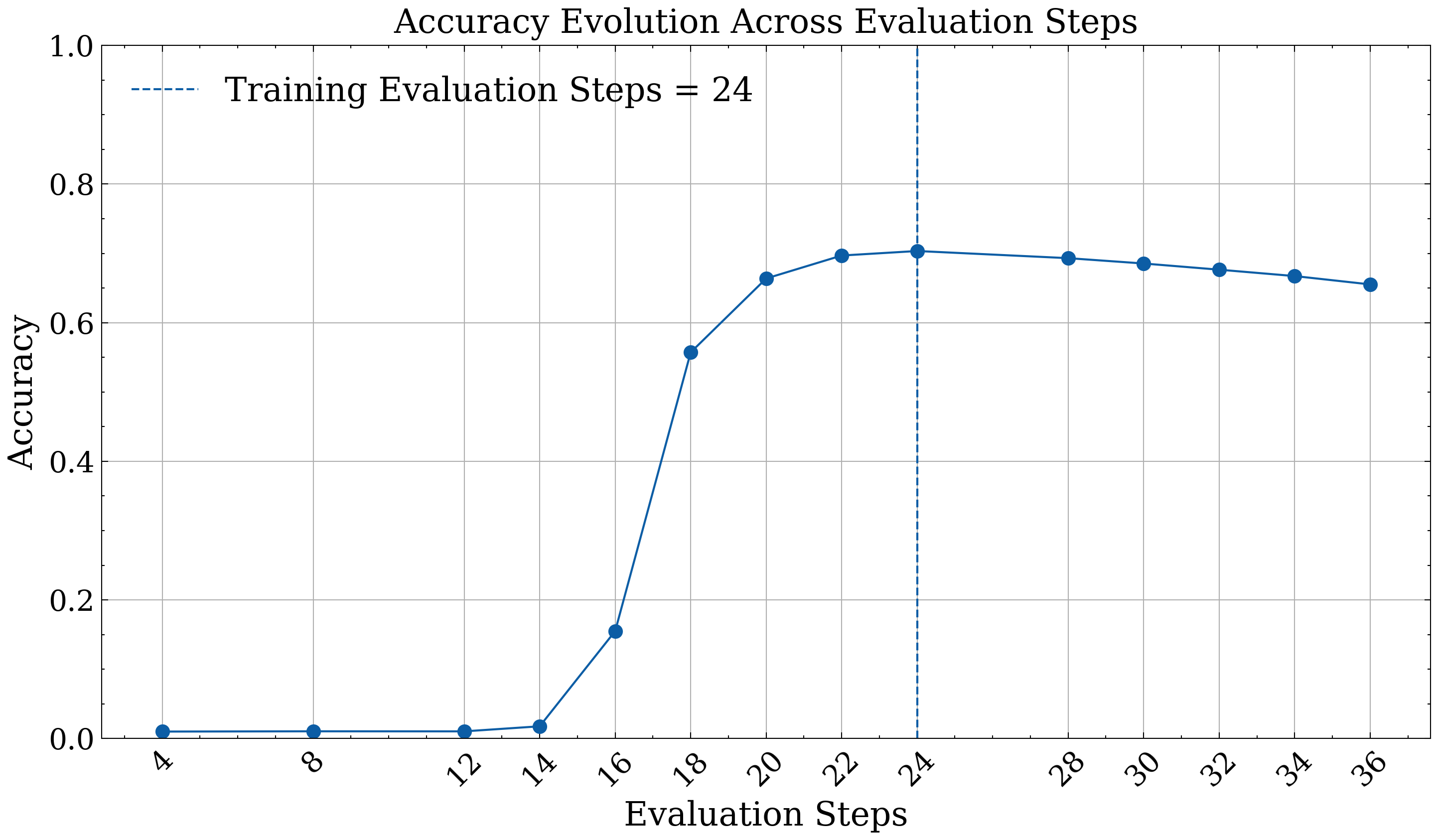}
&
\includegraphics[width=0.95\columnwidth]{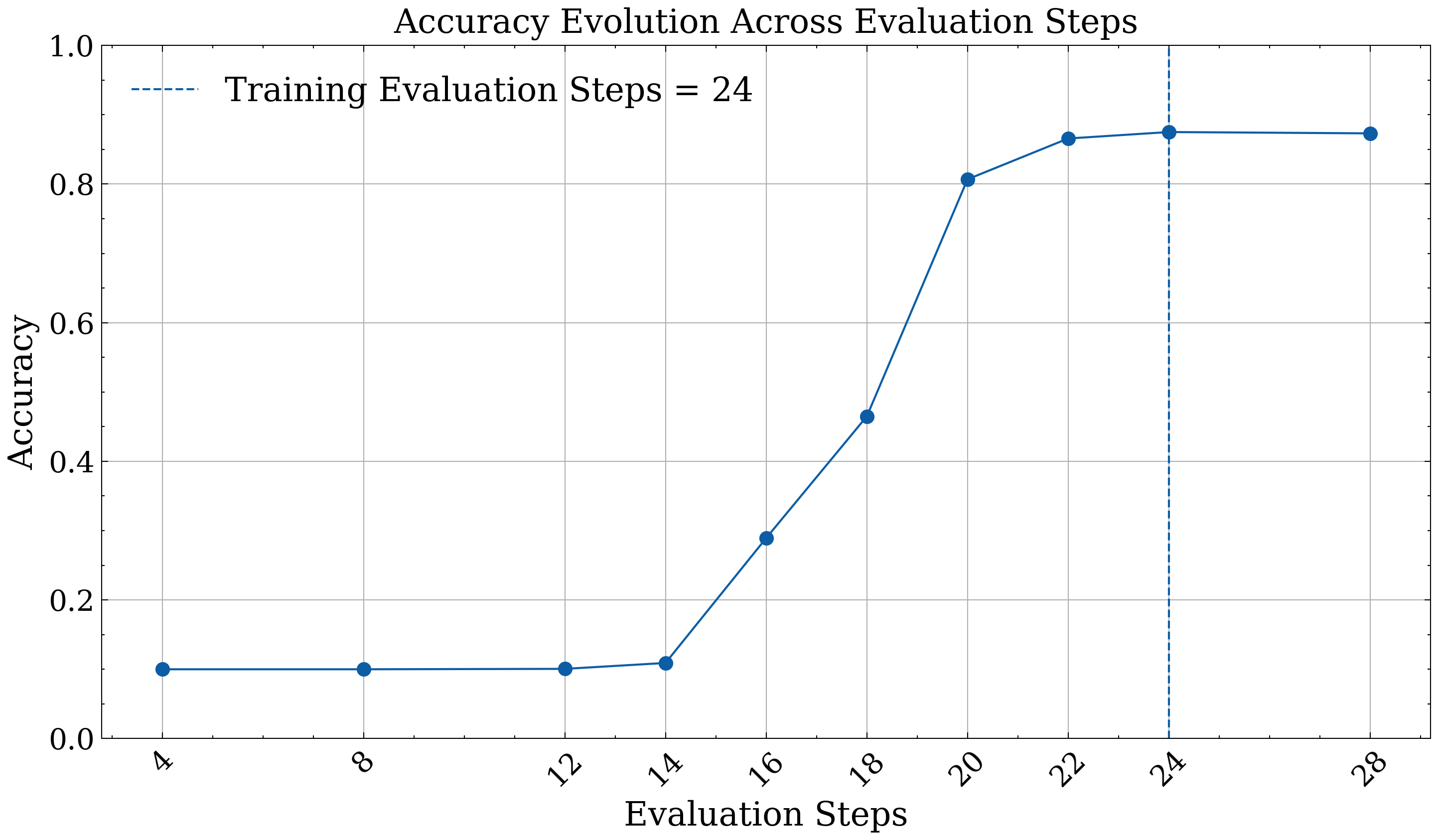}
\end{tabular}
    \caption{Line-plot with different evaluation steps on, from left to right, CIFAR-100 and CIFAR-10. It can be seen that it is suitable to stop the inference with 6 steps fewer. The decrease in performance is negligible with respect to the results using the same evaluation steps as in training.}
\label{fig: Eval Steps}
\end{figure*}

The hyperparameters used in each of the best results of the experiments can be seen in the \cref{tab: Hyperparameters}. The MLP ratio belongs to the MLP that follows the same structure as the ViT, where an intermediate layer is used. Playing with the intermediate layer increases and reduces the parameters significantly. The radius defined in the JasMin loss \eqref{eq:JaSMin} is set to 10

\section{Extra Experiments}

\begin{figure}[t]
    \centering
    \includegraphics[width=0.6\columnwidth]{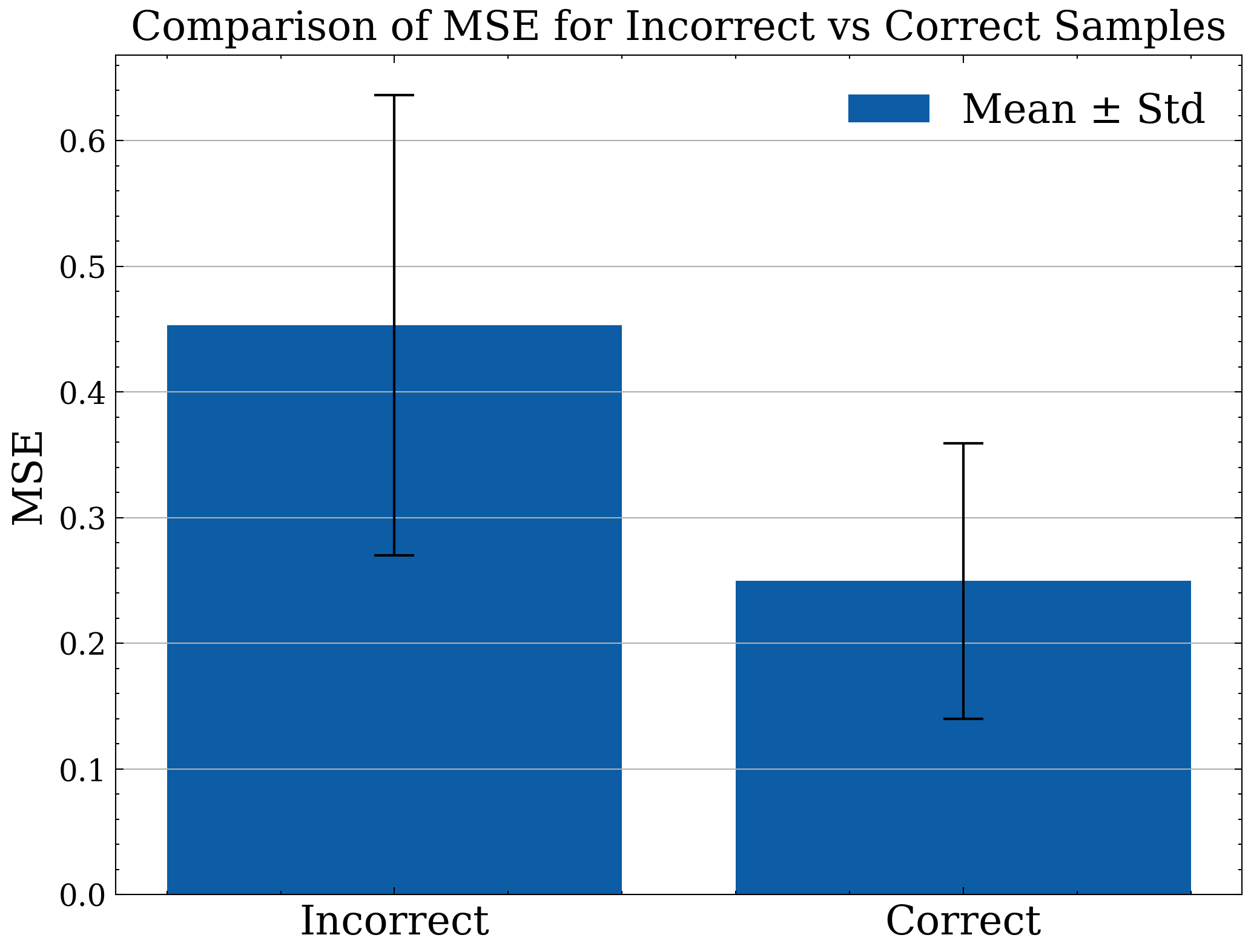}
    \caption{Barplot showing that the cases where the CLS token of the \ode is above a certain distance, the classification is badly done; however, when it is below, it classifies correctly as the teacher. }
    \label{fig: MSE Distances}
\end{figure}

In the paper we claim that we could quantify when the model can stop evaluating $\psi$ as it reaches the contraction region of the classification head. First in \cref{fig: MSE Distances}, we show that the distance to the CLS token of the teacher is correlated with the correct and the incorrect samples, as in \cref{fig: contraction results}, but the plot zooms in on the case when the teacher is correct.

\begin{figure}
    \centering
    \includegraphics[width=1.1\columnwidth]{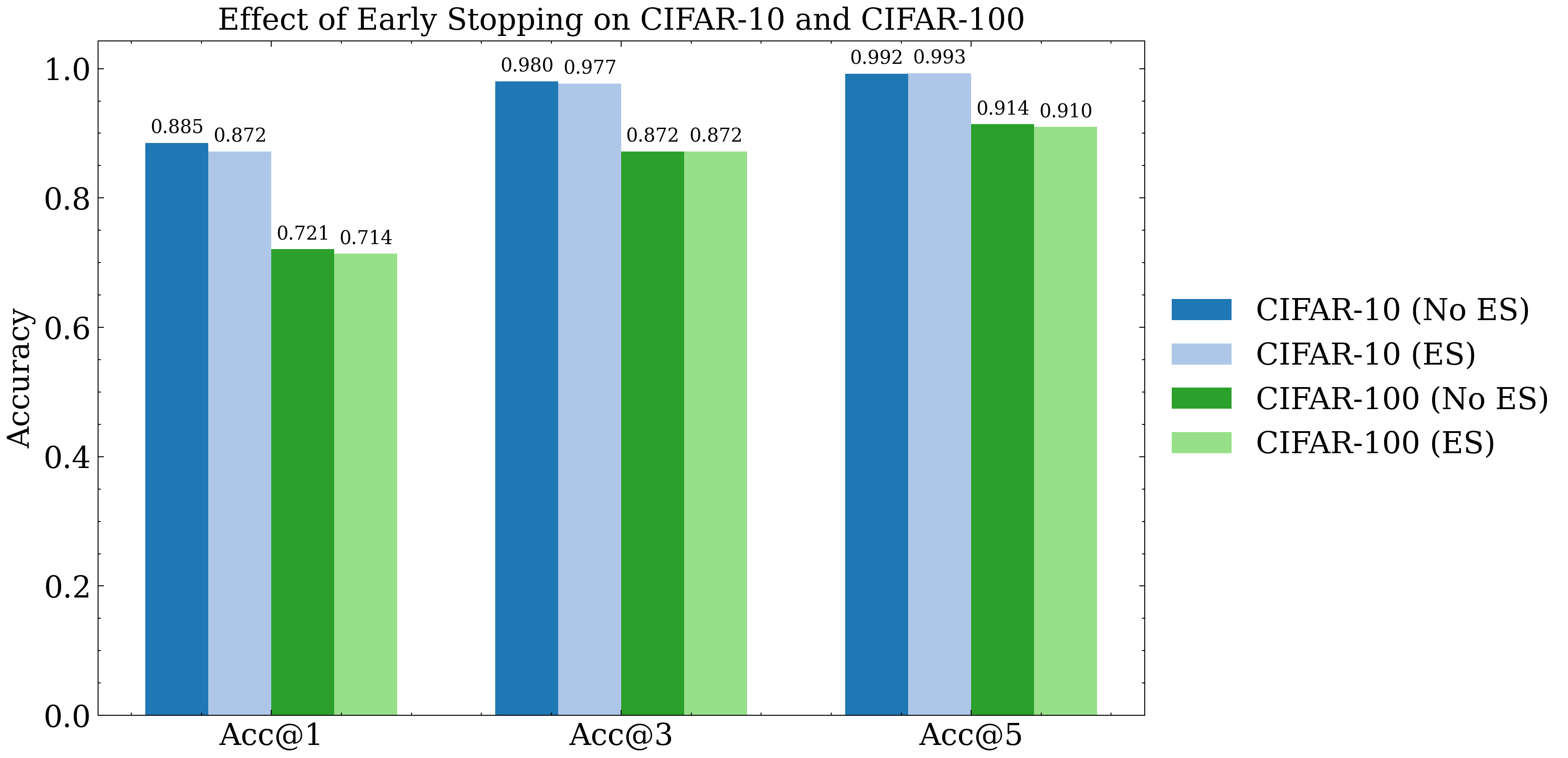}
    \caption{Bar-plot with the comparison applying the Early Stopping. ES means that when each of the losses is below \eqref{eq: Error desglossed} with a patience of 10 epochs, the model stops training.}
    \label{fig: ES}
\end{figure}

Next, we experiment to validate the hypothesis by varying the number of evaluation steps and the model's horizon. In \cref{fig: Eval Steps} and  \cref{fig: horizon evaluations}, the different results are depicted. The interpretation of the plots follows the behavior described in \cite{ott2021resnet}. We trained the model with a specific resolution of steps and a particular horizon. At test time, the increase or decrease in the number of steps is bounded to a certain range where the results are almost the same. It means that up to a certain evaluation, the representation has reached the contractive region where the classification can be performed without further penalization.

Finally, we experimented with using \eqref{eq:approx_error} as an early stopping criterion in the teacher-student framework. In this case, the Lipschitz constant is set to 0.5 (L=0.5) as described in \cite{yudin2025payattentionattentiondistribution}. For the computation of $\| \ddot{x}\|_\infty$, let \(W_Q, W_K, W_V \in \mathbb{R}^{D \times d}\) denote the query, key, and value projection matrices.  
Given a radius \(R > 0\), a Lipschitz constant \(L > 0\), and  \(N = \texttt{num\_eval\_steps}\), the upper bound computed in the code can be written as:
\begin{equation}\label{eq: Error desglossed}
err = 
\frac{e^{L} - 1}{2 L N} \,
\frac{
R^{2}\,\|W_V\|_{2}\,
\left( R\,\| W_K W_Q^{\top} \|_{2} + \sqrt{d} \right)
}{
N^{2}\sqrt{d}
}.    
\end{equation}

\begin{figure}
    \centering
    \includegraphics[width=\columnwidth]{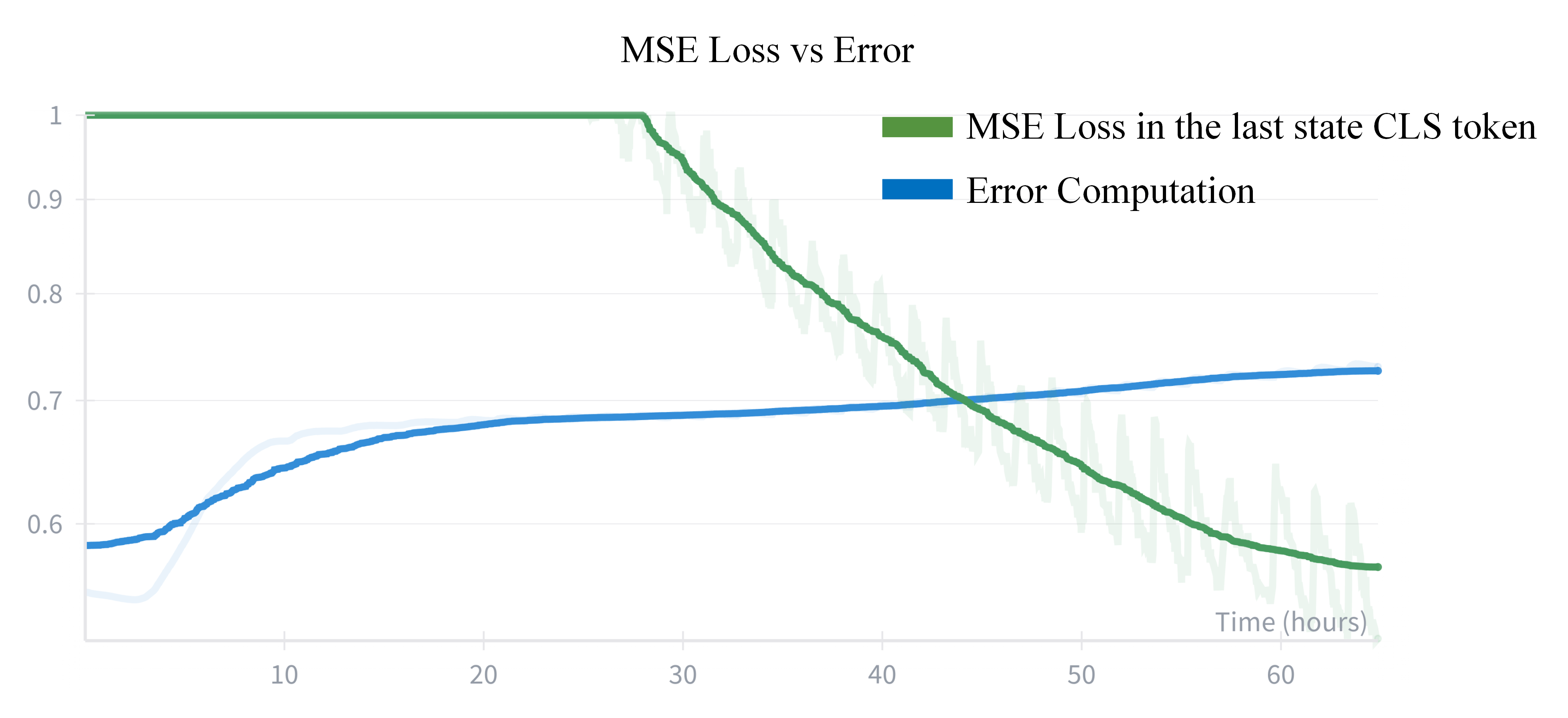}
    \caption{Line-plot showing the tracking of the early stopping. In this case, when the running average of the last samples of the MSE loss is below the upper bound error, the training starts to plateau as discussed before.}
    \label{fig Crossing Error}
\end{figure}

Here, $R$ is set to a radius of 10, as well as the JasMin loss. The effect of applying early stopping versus not using it is shown in \cref{fig: ES}. This upper-bound–based quantification of the error provides a principled criterion to determine when to stop training. Since the upper bound depends on the trainable parameters, it tends to increase during training; therefore, we monitor the point at which both the losses and the upper bound intersect. In \cref{fig Crossing Error} there is an example of the tracking of the bounding error and the MSE loss in the last state of the \ode trajectory and the last hidden representation of the ViT. Using early stopping avoided almost 15 hours of computation.

\end{document}